\documentclass[10pt,twocolumn,letterpaper]{article}

\usepackage{cvpr}              

\usepackage{CJK}
\usepackage{multicol}
\usepackage{color}
\usepackage{xcolor}
\usepackage{colortbl}
\usepackage{bm}
\usepackage{extarrows}
\usepackage{float}
\usepackage{bbm}
\usepackage{booktabs}
\usepackage{multirow}
\usepackage{amsmath}
\usepackage{amssymb}
\usepackage{algorithm}
\usepackage{algorithmic}
\usepackage{makecell}
\usepackage{amsmath}
\usepackage{amssymb}
\usepackage{amsthm}
\usepackage[utf8]{inputenc}
\usepackage{amsfonts} 
\usepackage{caption} 
\usepackage{newfloat}
\usepackage{listings}

\usepackage{xcolor}
\definecolor{thmcolor}{HTML}{0201F5}

\newtheoremstyle{coloredthm}
  {3pt}
  {3pt}
  {\itshape}
  {}
  {\bfseries\color{thmcolor}}
  {.}
  {.5em}
  {}

\theoremstyle{coloredthm}

\definecolor{cvprblue}{rgb}{0.21,0.49,0.74}
\definecolor{supplcolor}{HTML}{A32D26}

\definecolor{noise20color}{HTML}{FFF9E6} 
\definecolor{noise50color}{HTML}{EDF7ED} 
\definecolor{noise80color}{HTML}{EBF5FB} 

\usepackage[pagebackref,breaklinks,colorlinks,citecolor=brown]{hyperref}
\usepackage{thmtools}

\def\paperID{*****} 
\def\confName{CVPR}
\def\confYear{2026}

\title{$\mathbb{R}^3$: Composed Video Retrieval via Reasoning-Guided Recalling and Re-ranking}

\author{Zixu Li$^{1}$~~~~~Yupeng Hu$^{1}$~~~~~Zhiheng Fu$^1$~~~~~Zhiwei Chen$^1$~~~~~Weili Guan$^{2}$~~~~~Liqiang Nie$^2$ \vspace{2mm}\\
$^1$Shandong University\hspace{1.5cm}$^2$Harbin Institute of Technology (Shenzhen)\hspace{1.5cm}\\
{\tt\small \{lizixu.cs, fuzhiheng8, zivczw, honeyguan, nieliqiang\}@gmail.com;} \\ 
{\tt\small huyupeng@sdu.edu.cn}\\
}

\begin{document}
\maketitle
\begin{abstract}
The CoVR-R challenge evaluates composed video retrieval, where a system must retrieve a target video from a large gallery given a reference video and a textual edit instruction. This setting is not a standard video-text retrieval problem: the query is defined by both the visual evidence in the source video and the transformation implied by the edit. A strong embedding model can provide scalable candidate recall, but it may under-express target-side consequences such as state changes, action replacement, object preservation, or temporal consistency. 
A pairwise multimodal reranker can verify such details more directly, but exhaustive reranking over the full gallery is computationally infeasible.
We present $\mathbb{R}^3$, a zero-shot composed video retrieval pipeline built around \emph{\textbf{R}easoning}-guided \emph{\textbf{R}ecalling} and \emph{\textbf{R}eranking}. The core idea is to turn the source-edit query into a reasoning-grounded retrieval program rather than treating the edit text as a short caption. First, the model generates a reasoning trace that describes the expected target video after applying the edit. Then the trace is encoded together with the source video as a reasoning-augmented query, and its retrieval score is fused with the base composed query through an agreement-gated residual rule. At last, a re-ranker verifies the recalled candidates with direct source-candidate comparison. 
Experiments have demonstrated the effectiveness of our method in addressing this challenge. Codes are available on \href{https://github.com/Lee-zixu/R-3}{https://github.com/Lee-zixu/R-3}.
\end{abstract}
    
\section{Introduction}
\label{sec:intro}

Composed video retrieval~\cite{HUD,covr,REFINE,covr-2,ReTrack} aims to retrieve a target video from a gallery given a reference video and a natural language modification. Following composed image retrieval~\cite{ConeSep,Air-Know,sprc,ENCODER,OFFSET,PAIR,MEDIAN}, unlike standard video-text retrieval~\cite{shen2025tempme,TempRet}, the query is not a complete caption, but a modification text that conveys the users' more precise requirements. It is defined by a visual anchor and a transformation: the source video specifies the objects, actions, scene, camera view, and temporal context that should be preserved, while the edit instruction specifies how the desired target should differ. A correct system must therefore solve both preservation and change, which makes CoVR-R a compositional retrieval benchmark rather than a conventional semantic matching task. Furthermore, as a fundamental task in multi-modal interaction, CVR supports real-world applications such as multimodal reasoning~\cite{qwenvl,ERASE,EgoAdapt}, multimodal learning~\cite{EgoAction,qwen25vl,OmniEgo-R}, and cross-modal retrieval~\cite{fang2023uatvr,STABLE,egovlpv2}.

A common solution is embedding-based retrieval~\cite{FineCIR,INTENT,HABIT,TEMA}. The reference video and edit text are encoded into a query embedding, gallery videos are encoded into target embeddings, and candidates are ranked by cosine similarity. This design is efficient and naturally supports large galleries, but it has an important limitation: the modification is often compressed into a short text condition. Many edits imply target-side consequences that are not explicitly named. For example, changing an action may imply a different object state, changing a scene may require preserving the actor and motion, and modifying camera framing may require distinguishing foreground changes from background context. A single composed embedding can recall plausible candidates, but it may not explicitly represent these implied visual consequences.

Pairwise multimodal reranking addresses a complementary limitation~\cite{MELT,HINT}. By directly comparing the source video, a candidate target video, and the edit instruction, a reranker can verify whether the candidate satisfies the edit while preserving unchanged context. However, this verification is computationally expensive because it requires processing two videos for each candidate pair. Exhaustive reranking over the full gallery is therefore infeasible. In practice, reranking can only be used after a high-recall retrieval stage has already reduced the search space.

The composed video retrieval field further suggests that useful predictions are not only ranked lists but also explanations of the intended target~\cite{covr-r}. A reasoning trace can make implicit edit effects explicit, but a naive use of reasoning has a risk. If it is used as an unconstrained replacement query, it may over-specify details that are not visually supported by the source video or instruction. Thus, reasoning must be integrated into retrieval in a controlled manner.

These observations lead to two challenges. \textbf{C1: implicit reasoning and reliability.} The target video may contain implicit state, action, or context changes not explicitly stated in the edit text. While generating visual reasoning traces can help decode these nuances, integrating such reasoning without proper safeguards risks introducing noise or over-specific biases that degrade retrieval.
\textbf{C2: context preservation and verification efficiency.} Retrieval must strike a delicate balance between preserving source-side context and capturing target modifications. Moreover, while coarse embedding-based search scales efficiently to maintain this balance, relying solely on it lacks precision, whereas deploying accurate multimodal pairwise verification across all candidates is computationally prohibitive.

To address these challenges, we propose $\mathbb{R}^3$, a zero-shot pipeline organized around \emph{Reasoning}, \emph{Recalling}, and \emph{Re-ranking}. The reasoning uses Qwen3-VL-Thinking to generate a target-oriented reasoning trace before candidate recall. The recalling encodes both the base source-edit query and the reasoning-augmented query, then fuses their scores through an agreement-gated residual rule. The re-ranking applies Qwen3-VL-Reranker only to the recalled candidates, using pairwise verification to improve local ordering. 
In this way, reasoning becomes a recalling signal, while the gate and residual weight prevent it from overriding the original composed query.

Our contributions are summarized as follows:
\begin{itemize}
    \item We formulate this challenge as a reasoning-guided coarse-to-fine retrieval problem, identifying the limitations of existing approches.
    \item We introduce the reasoning trace before retrieval and uses it as a controlled query expansion signal.
    \item We design an agreement-gated residual fusion rule that lets reasoning adjust embedding scores.
\end{itemize}

\begin{figure*}[ht]
	\includegraphics[width=0.98\linewidth]{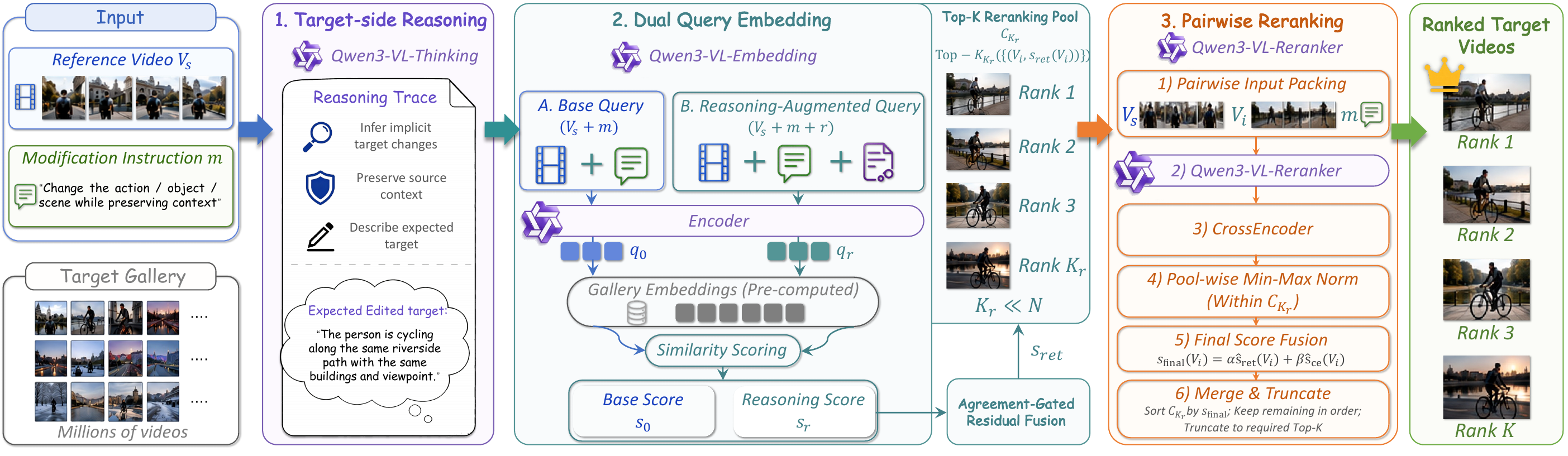}
	\caption{Overview of the proposed $\mathbb{R}^3$ framework for composed video retrieval. Given a reference video $V_s$, a modification instruction $m$, and a target gallery, $\mathbb{R}^3$ generates a reasoning trace $r$ to expose implicit edit consequences and preservation constraints, then performs dual-query retrieval using both the base query $(V_s,m)$ and the reasoning-augmented query $(V_s,m,r)$. Their scores are combined through agreement-gated residual fusion to obtain $s_{\mathrm{ret}}$, from which a compact Top-$K_r$ re-ranking pool is selected. A re-ranker further verifies each source-candidate pair, and the normalized retrieval and re-ranking scores are fused to produce the final Top-$K$ ranked target videos.}
	\label{fig:Framework}
\end{figure*}

\section{Methodology}
\label{sec:method}

$\mathbb{R}^3$ is a unified inference-time pipeline built on frozen Qwen3-VL components. The method is designed around the challenges discussed above. Instead of solving these issues with task-specific training, $\mathbb{R}^3$ turns each query into a structured \textbf{R}easoning--\textbf{R}ecalling--\textbf{R}e-ranking program. Qwen3-VL-Thinking first describes the expected target, Qwen3-VL-Embedding performs scalable reasoning-guided recall, and Qwen3-VL-Reranker verifies the recalled candidates. 
In the following, we first explain the design rationale, then formulate the problem and describe each module.

\subsection{Design Rationale}
\label{sec:rationale}

The design follows a coarse-to-fine principle. First, the system must understand the intended edited target before searching the gallery. This motivates a reasoning stage that translates the source video and edit instruction into an explicit target-side description. Second, the reasoning trace affects retrieval. We use a reasoning-augmented query encoding. Third, the generated trace should not replace the original query, because it may contain unsupported details. This motivates the agreement gate and residual score fusion. Fourth, final candidate ordering should be decided by direct source-candidate comparison.

Thus, reasoning trace generation addresses implicit target transformation. Base and reasoning-augmented retrieval address the preservation-change balance. Agreement-gated residual fusion addresses reasoning reliability. Candidate reranking addresses fine-grained pairwise verification while avoiding exhaustive gallery scoring. 

\subsection{Problem Formulation and Overview}
\label{sec:problem}

For each sample, the input is a source video $V_s$, a modification instruction $m$, and a gallery of candidate target videos $\mathcal{G}=\{V_i\}_{i=1}^{N}$. The goal is to output a ranked list of target video ids,
\begin{equation}
    \hat{\mathcal{Y}} =
    \operatorname{Rank}(V_s,m,\mathcal{G}),
\end{equation}
where high-ranked candidates should match the visual result of applying $m$ to $V_s$. $\mathbb{R}^3$ decomposes the prediction as
\begin{equation}
    \hat{\mathcal{Y}} =
    \mathcal{R}_{\text{rank}}
    \circ
    \mathcal{R}_{\text{ret}}
    \circ
    \mathcal{R}_{\text{think}}
    (V_s,m,\mathcal{G}),
\end{equation}
where $\mathcal{R}_{\text{think}}$ generates a reasoning trace, $\mathcal{R}_{\text{ret}}$ performs reasoning-guided embedding retrieval, and $\mathcal{R}_{\text{rank}}$ applies pairwise reranking to the recalled candidates. This decomposition is a test-time program rather than a trainable network.

The important design choice is that reasoning is placed before retrieval. The generated trace is not only an interpretable output field; it is used to construct an additional query representation that can influence candidate recall. At the same time, the influence is guarded by an agreement score and a residual weight, so that the original source-edit query remains the main retrieval anchor.

\subsection{Gallery Evidence Encoding}
\label{sec:gallery}

$\mathbb{R}^3$ first encodes the gallery with Qwen3-VL-Embedding. For each candidate video $V_i$, we form a video-only document item and apply the document instruction that asks the model to represent visible objects, actions, state changes, scene context, camera framing, and temporal progression. 
The candidate embedding is
\begin{equation}
    \mathbf{d}_i =
    \operatorname{Norm}
    \big(f_{\text{emb}}(V_i; I_{\text{doc}})\big),
\end{equation}
where $f_{\text{emb}}$ is the Qwen3-VL-Embedding encoder and $\operatorname{Norm}$ denotes $L_2$ normalization. All candidate embeddings are stored in a matrix: 
\begin{equation}
    \mathbf{D}=[\mathbf{d}_1,\ldots,\mathbf{d}_N]^\top .
\end{equation}
This stage is executed once for each gallery split. At inference time, retrieval is therefore reduced to query encoding and matrix multiplication, which keeps the expensive video-document encoding outside the online loop.

\subsection{Reasoning Trace Generation}
\label{sec:reasoning}

The edit instruction alone may not state all properties that the target video should contain. We therefore introduce an inference-time reasoning module. Given $(V_s,m)$, Qwen3-VL-Thinking produces a target-oriented reasoning trace
\begin{equation}
    r =
    \Pi_{\text{final}}
    \big(f_{\text{think}}(V_s,m; P_{\text{reason}})\big),
\end{equation}
where $P_{\text{reason}}$ is the reasoning prompt and $\Pi_{\text{final}}$ extracts the final external answer from the thinking-model output. The trace should explain the expected visual transformation and preservation constraints.

The prompt encourages a complete target-side paragraph for edits involving object replacement, action change, state transition, camera/framing shift, motion change, or contextual preservation. Simple edits are allowed to remain concise. Thus, the reasoning trace records what the target is expected to show, which visual context should remain unchanged, and which details should be checked during retrieval and reranking.

\subsection{Prompting and Interface Design}
\label{sec:prompting}
All modules are connected through lightweight prompts and structured model interfaces. The embedding document prompt asks Qwen3-VL-Embedding to represent each candidate video by visible objects, actions, state changes, scene context, camera framing, and temporal progression. The retrieval prompt asks the same encoder to retrieve the target video that matches the reference video after applying the edit instruction. This keeps gallery encoding and query encoding aligned around the same composed-video objective.

The reasoning prompt has a different role. It asks Qwen3-VL-Thinking to infer the expected target video and output a concise external reasoning paragraph. The final output is not a dialogue response and not an internal chain-of-thought transcript; it is a benchmark-facing description of the visual transformation and preserved context. The reranker prompt then asks Qwen3-VL-Reranker to score whether the candidate target video is a good match for the edited source video. These prompts define the interfaces between the three frozen models and make the system reproducible without training additional parameters.

\subsection{Base and Reasoning-Augmented Retrieval}
\label{sec:retrieval}

The base composed query uses the source video and the edit instruction:
\begin{equation}
    T_0(m)=
    \text{``Reference video after edit instruction: }m\text{''}.
\end{equation}
It is encoded as: 
\begin{equation}
    \mathbf{q}_0 =
    \operatorname{Norm}
    \big(f_{\text{emb}}(V_s,T_0(m); I_{\text{ret}})\big).
\end{equation}
The base retrieval score for candidate $V_i$ is the cosine score, as follows, 
\begin{equation}
    s_0(V_i)=\mathbf{d}_i^\top \mathbf{q}_0 .
\end{equation}

To connect the reasoning trace with retrieval, we construct a second query view,
\begin{equation}
\begin{aligned}
    T_r(m,r) = & \text{``Reference video after edit instruction: }m \\
               & \text{. Reasoned target video description: }r\text{''},
\end{aligned}
\end{equation}
and encode it as: 
\begin{equation}
    \mathbf{q}_r =
    \operatorname{Norm}
    \big(f_{\text{emb}}(V_s,T_r(m,r); I_{\text{ret}})\big).
\end{equation}
The corresponding reasoning retrieval score is: 
\begin{equation}
    s_r(V_i)=\mathbf{d}_i^\top \mathbf{q}_r .
\end{equation}
The base query preserves the literal source-edit condition, while the reasoning-augmented query exposes inferred target-side consequences. Both scores are computed against the same precomputed gallery matrix.

\subsection{Agreement-Gated Residual Fusion}
\label{sec:gated}

Reasoning should help retrieval only when it remains consistent with the original query. We therefore compute an embedding-space agreement score, formulated as follows, 
\begin{equation}
    a = \mathbf{q}_0^\top \mathbf{q}_r .
\end{equation}
If $a<\tau$, where $\tau$ is a configurable threshold, the reasoning branch is blocked and the system ranks candidates by $s_0$ only. If the gate is satisfied, the reasoning score is applied as a residual correction:
\begin{equation}
    s_{\text{ret}}(V_i)
    =
    s_0(V_i)
    +
    \lambda\big(s_r(V_i)-s_0(V_i)\big).
\end{equation}
The weight $\lambda$ is intentionally small. It is set to $0.5$, so reasoning can adjust the candidate ordering but cannot replace the base composed query. Source-video aliases are assigned $-\infty$ before ranking to prevent degenerate self-retrieval. The top $K_r$ candidates under $s_{\text{ret}}$ are then passed to the reranking stage.

This module is the main connection between reasoning and search. It gives the generated trace a concrete scoring role, while the agreement gate and residual form prevent unsupported reasoning details from dominating candidate recall.

\subsection{Candidate Reranking}
\label{sec:reranking}

Embedding retrieval scores each candidate independently with respect to the query embedding. To verify fine-grained pairwise compatibility, $\mathbb{R}^3$ applies Qwen3-VL-Reranker to the recalled candidates. For each candidate $V_i$ in the top-$K_r$ set, the reranker receives the source video, candidate target video, and edit instruction. It predicts a scalar compatibility score:
\begin{equation}
    s_{\text{ce}}(V_i)=
    g_{\text{rank}}(V_s,V_i,m),
\end{equation}
where $g_{\text{rank}}$ denotes the Qwen3-VL cross-encoder reranker. The implementation uses the SentenceTransformers CrossEncoder.

More explicitly, let $\mathcal{C}_{K_r}$ be the reranking pool selected from embedding retrieval:
\begin{equation}
    \mathcal{C}_{K_r}
    =
    \operatorname{TopK}_{K_r}
    \big(\{(V_i,s_{\text{ret}}(V_i))\}_{i=1}^{N}\big).
\end{equation}
For each $V_i\in\mathcal{C}_{K_r}$, we build a cross-encoder input pair and score it:
\begin{equation}
    s_{\text{ce}}(V_i)
    =
    \text{CrossEncoder}
    \big(
        \Phi(V_s, m; I_{\text{rank}}),\;
        \Phi(V_i; I_{\text{rank}})
    \big),
\end{equation}
where $\Phi$ packs video and text inputs into the multimodal format expected by Qwen3-VL, and $I_{\text{rank}}$ is the reranker instruction. The CrossEncoder jointly encodes the source-candidate pair and produces a scalar compatibility score. A larger $s_{\text{ce}}$ means that the candidate is more likely to be a valid target for the edited source video.

For score composition, retrieval and reranking scores are min-max normalized inside the valid reranked set. Let: 
\begin{equation}
    m_z=\min_{V_j\in\mathcal{C}_{K_r}} s_z(V_j),\quad
    M_z=\max_{V_j\in\mathcal{C}_{K_r}} s_z(V_j),
\end{equation}
where $z\in\{\text{ret},\text{ce}\}$. The normalized score is: 
\begin{equation}
    \hat{s}_{z}(V_i)=
    \frac{s_z(V_i)-m_z}{M_z-m_z+\epsilon}.
\end{equation}
The final candidate score is
\begin{equation}
    s_{\text{final}}(V_i)=
    \alpha \hat{s}_{\text{ret}}(V_i)
    +
    \beta \hat{s}_{\text{ce}}(V_i).
\end{equation}
Thus, embedding retrieval remains the coarse similarity anchor, while re-ranking provides a pairwise verification correction for local ordering. This preserves the two-stage structure: retrieval decides which candidates are plausible enough to inspect, and re-ranking adjusts their order.

\section{Experiments}
\label{sec:experiments}

\subsection{Benchmark, Splits, and Submission Format}
\label{sec:setup}

We evaluates composed video retrieval over multiple video sources, including WebVid videos and Something-Something-V2 videos. Each query consists of a source video id and a textual modification. The system returns a ranked list of target video ids and a reasoning trace when required by testing phase. The final submission keeps the top 50 predictions per query. We report results on both validation and test splits, output from the official server.

    \begin{figure*}[t]
    \centering
    \small
    
    \begin{minipage}[t]{0.32\textwidth}
    \textbf{\small Stage 1: Embedding}\\[3pt]
    \setlength{\fboxsep}{6pt}%
    \fbox{\parbox{\dimexpr\linewidth-2\fboxsep-2\fboxrule}{
        \textit{\small Embedding instruction:}\\[2pt]
        \texttt{Retrieve the target video that matches the reference video after applying the edit instruction. The correct target should satisfy the requested visual change while preserving unchanged visual context, including objects, actions, state changes, scene, camera framing, and temporal progression.}
    }}
    \end{minipage}
    \hfill
    \begin{minipage}[t]{0.32\textwidth}
    \textbf{\small Stage 2: Recalling}\\[3pt]
    \setlength{\fboxsep}{6pt}%
    \fbox{\parbox{\dimexpr\linewidth-2\fboxsep-2\fboxrule}{
    \textit{\small Query input:}\\[2pt]
    \texttt{Reference video after edit instruction: \{modification\_text\}}\\[4pt]
    \textit{\small Reasoning-augmented query:}\\[2pt]
    \texttt{Reference video after edit instruction: \{modification\_text\}}\\
    \texttt{Reasoned target video description: \{reasoning\}}\\[4pt]
    }}
    \end{minipage}
    \hfill
    \begin{minipage}[t]{0.32\textwidth}
    \textbf{\small Stage 3: Re-ranking}\\[3pt]
    \setlength{\fboxsep}{6pt}%
    \fbox{\parbox{\dimexpr\linewidth-2\fboxsep-2\fboxrule}{
    \textit{\small Reranker instruction:}\\[2pt]
    \texttt{Score how well the candidate target video matches the reference video after applying the edit instruction. A good candidate satisfies the requested visual change, preserves unchanged context, and remains temporally and visually consistent with the intended edited result.}
    }}
    \end{minipage}
    
    \caption{\textbf{Prompts used across the three stages of our pipeline.} Template variables \texttt{\{modification\_text\}} and \texttt{\{reasoning\}} are filled per sample at runtime.}
    \label{fig:prompts}
    \end{figure*}

\begin{table*}[t]
\centering
\caption{Ablation results of $\mathbb{R}^3$. All scores are evaluated at the official server.    {}$^*$Baseline is frozen Qwen3-VL-Embedding-8B. We build on this model.}
\label{tab:ablation-template}
\begin{tabular}{@{}cllccccc@{}}
\Xhline{1pt}
\textbf{Setting} &\textbf{Method} &\textbf{Model}& \textbf{R@1} & \textbf{R@5} & \textbf{R@10} & \textbf{R@50}  \\
\hline
\hline
A-0&Baseline$^{*}$ & Qwen3-VL-Embedding-8B  &{94.78} & \underline{99.66} & \underline{99.83} & \textbf{100.00}  \\
A-1&+ Reasoning &Qwen3-VL-32B-Thinking  &\underline{95.12} & \underline{99.66} & \underline{99.83} & \textbf{100.00}  \\
\hline
A-2&+ Re-ranking &Qwen3-VL-Reranker-8B  &\textbf{98.82} & \textbf{100.00} & \textbf{100.00} & \textbf{100.00}  \\

\Xhline{1pt}
\end{tabular}
\end{table*}

\label{sec:performance}
\begin{table}[h]
\centering
\caption{Main CoVR-R retrieval performance on validation and test splits. All scores are evaluated at the official server.}
\label{tab:main-results}
\begin{tabular}{@{}lcccc@{}}
\Xhline{1pt}
\textbf{Split} & \textbf{R@1} & \textbf{R@5} & \textbf{R@10} & \textbf{R@50}  \\
\hline
Validation & 95.44 & 99.49 & 99.77 & 99.96  \\
Test & 98.82 & 100.00 & 100.00 & 100.00 \\ 
\Xhline{1pt}
\end{tabular}
\end{table}

\subsection{Implementation Details}
\label{sec:implementation}

Figure~\ref{fig:prompts} presents the prompts used across the three stages of our pipeline.
No task-specific training or fine-tuning is performed. All models are frozen. Gallery videos are encoded once with Qwen3-VL-Embedding-8B and saved as compressed NumPy stores. The reasoning trace is outputed from Qwen3-VL-32B-Thinking (temperature = $0.2$, top-p = $0.9$ and a maximum of 4096 tokens). The recalling process enables reasoning-guided scoring with the gated residual mode, and the main residual weight is $\lambda=0.5$. $\tau$ is set to $0.5$. And we set $\alpha=0.4$ and $\beta=0.6$.
The re-ranking process is performed by Qwen3-VL-Reranker-8B, with Top-K set to 50 and the internal CrossEncoder used.
The re-ranking process is performed by Qwen3-VL-Reranker-8B via the SentenceTransformers CrossEncoder interface, with Top-$K_r$ set to $50$. The re-ranker instruction asks the model to score how well the candidate target video matches the edited source video, considering both the requested visual change and preserved context. For score fusion, we use linear combination with weights $\alpha=0.4$ (retrieval) and $\beta=0.6$ (reranker). All experiments are performed on a single node of 8 RTXPRO6000 GPUs.

\subsection{Ablation Studies}
\label{sec:ablation}

Table~\ref{tab:ablation-template} summarizes the ablation protocol. The first row measures the direct embedding recall baseline. The second row isolates re-ranking. The final row corresponds to the full $\mathbb{R}^3$ pipeline. These results are intended to show whether reasoning improves recalling before re-ranking and whether re-ranking improves local ordering after recall.

Comparing A-0 and A-1, reasoning-augmented recalling provides a modest improvement of +$0.34$ R@1, with R@50 already saturated at $100.00$ for both settings. This suggests that the base embedding retrieval already captures most of the edit semantics, and the reasoning trace contributes marginal recall gains..

Comparing A-1 and A-2, re-ranking delivers a substantial improvement of +$3.70$ R@1, confirming that local reordering is the primary driver of top-1 accuracy. The reranker performs direct pairwise verification between the source video and candidate, which is more precise than embedding cosine similarity for distinguishing visually similar targets. Combined with reasoning, the full $\mathbb{R}^3$ pipeline achieves $98.82$ R@1 on validation and $100.00$ R@5 and above, demonstrating that the coarse-to-fine design is effective.

\subsection{Validation and Test Performance}
Table~\ref{tab:main-results} provides the main results for the official validation and test splits. On the validation split, the system achieves $95.44$ R@1 and $99.96$ R@50. On the test split, R@5 through R@50 reach $100.00$, with R@1 at $98.82$. These results indicate that embedding retrieval already provides strong recall and that reranking effectively reorders the top candidates.

\section{Observations and Insights}
\label{sec:observations}

The main observation is that reasoning and re-ranking address different errors. Reasoning-guided retrieval is most useful when the edit implies a target-side transformation that is not fully captured by the short modification text, such as state change, action replacement, or preservation of contextual details. Re-ranking is most useful when the correct target is already inside the recalled set but must be distinguished from visually similar candidates.

The agreement-gated residual rule is designed for this trade-off. It lets the reasoning query influence retrieval when it remains aligned with the base composed query, but falls back to the base embedding score when the reasoning branch appears inconsistent. This keeps the method robust while still giving the reasoning trace a functional role in candidate search.

\section{Conclusion}
\label{sec:conclusion}

We presented $\mathbb{R}^3$, a zero-shot composed video retrieval pipeline that integrates reasoning, recalling, and re-ranking into a single inference program. The method addresses the main limitations of direct embedding search and exhaustive re-ranking by generating a target-side reasoning trace before retrieval, using it through agreement-gated residual query fusion, and applying a reranker only to the recalled candidates. This design gives reasoning a functional role in search while keeping the original source-edit query as the anchor.

{
    \small
    \bibliographystyle{ieeenat_fullname}
    \bibliography{main}
}

\end{document}